\documentclass[journal]{IEEEtran}

\usepackage{colortbl,booktabs}
\usepackage{amsmath}
\usepackage{times}
\usepackage{epsfig}
\usepackage{graphicx}
\usepackage{amsmath}
\usepackage{amssymb}
\usepackage{color}
\usepackage{wrapfig}
\usepackage{multirow}
\usepackage{cite}
\usepackage{algorithm}
\usepackage{algorithmic}
\usepackage{subfigure}

\ifCLASSINFOpdf
\else
\fi
\begin{document}

\title{JigsawGAN: Auxiliary Learning for Solving Jigsaw Puzzles with Generative Adversarial Networks}

\author{Ru~Li,~\IEEEmembership{Student Member,~IEEE,}
        Shuaicheng~Liu,~\IEEEmembership{Member,~IEEE,}
        Guangfu~Wang,
        Guanghui~Liu,~\IEEEmembership{Senior Member,~IEEE,}
        Bing~Zeng,~\IEEEmembership{Fellow,~IEEE}
        
\thanks{
Manuscript received January 12, 2021; revised June 24, 2021 and September 21, 2021; accepted September 30, 2021.
Date of publication December 7, 2021; date of current version December 15, 2021. 
This work was supported in part by the National Natural Science Foundation of China (NSFC) under Grant 61872067, Grant 62071097, Grant 62031009, and Grant 61720106004.  in part by the ``111" Projects under Grant B17008, in part by Sichuan Science and Technåology Program under Grants 2019YFH0016. The associate editor coordinating the review of this manuscript and approving it for publication was Dr. Charles Kervrann. (\emph{Corresponding authors: Guanghui Liu and Shuaicheng Liu}.)

Ru~Li, Shuaicheng~Liu, Guanghui~Liu and Bing~Zeng are with School of Information and Communication Engineering, University of Electronic Science and Technology of China, Chengdu 611731, China  (e-mail: guanghuiliu@uestc.edu.cn; liushuaicheng@uestc.edu.cn)

Guangfu~Wang is with Megvii Technology, Chengdu 610000, China.

Digital Object Identifier 10.1109/TIP.2021.3120052
}
}


\maketitle

\begin{abstract}
The paper proposes a solution based on Generative Adversarial Network (GAN) for solving jigsaw puzzles. The problem assumes that an image is divided into equal square pieces, and asks to recover the image according to information provided by the pieces. Conventional jigsaw puzzle solvers often determine the relationships based on the boundaries of pieces, which ignore the important semantic information. In this paper, we propose JigsawGAN, a GAN-based auxiliary learning method for solving jigsaw puzzles with unpaired images (with no prior knowledge of the initial images). We design a multi-task pipeline that includes, (1) a classification branch to classify jigsaw permutations, and (2) a GAN branch to recover features to images in correct orders. The classification branch is constrained by the pseudo-labels generated according to the shuffled pieces. The GAN branch concentrates on the image semantic information, where the generator produces the natural images to fool the discriminator, while the discriminator distinguishes whether a given image belongs to the synthesized or the real target domain. These two branches are connected by a flow-based warp module that is applied to warp features to correct the order according to the classification results. The proposed method can solve jigsaw puzzles more efficiently by utilizing both semantic information and boundary information simultaneously. Qualitative and quantitative comparisons against several representative jigsaw puzzle solvers demonstrate the superiority of our method.
\end{abstract}


%
\IEEEpeerreviewmaketitle

\section{Introduction}\label{sec:introduction}
Solving the jigsaw puzzle is a challenging problem that involves research in computer science, mathematics and engineering. It abstracts a range of computational problems that a set of unordered fragments should be organized into their original combination. Numerous applications are subsequently raised, such as reassembling archaeological artifacts~\cite{papaodysseus2002contour,brown2008system,toler2010multi,pintus2016survey}, recovering shredded documents or photographs~\cite{marques2009reconstructing,liu2011automated,deever2012semi} and genome biology~\cite{marande2007mitochondrial}. Standard jigsaw puzzles are made by dividing images into interlocking patterns of pieces. With the pieces separated from each other and mixed randomly, the challenge is to reassemble these pieces into the original image. The degree of difficulty is determined by the number of pieces, the shape of the pieces and the graphical composition of the picture itself. The automatic solution of puzzles, without having any information on the underlying image, is NP-complete~\cite{altman1989solving,demaine2007jigsaw}. This task relies on computer vision algorithms, such as contour or feature detection~\cite{sivapriya2018automatic}. Recent development of deep learning (DL) opens bright perspectives for finding better reorganizations more efficiently. 

The first algorithm that attempted to automatically solve general puzzles was introduced by Freeman and Gardner~\cite{freeman1964apictorial}. The original approach was designed to solve puzzles with 9 pieces by only considering the geometric shape of the pieces. Various puzzle reconstruction methods are then introduced, which obey a basic operation that the unassigned piece should link to an existing piece by finding its best neighbor according to some affinity functions~\cite{chung1998jigsaw,papaodysseus2002contour,pomeranz2011fully} or matching contours~\cite{radack1982jigsaw,kong2001solving,liu2011automated}. These algorithms are slow because they keep iterating themselves until a proper solution is produced, as long as all parts are put in place. Moreover, the `best neighbors' may be false-positives due to the occasionally unobvious continuation between true neighbors. Some subsequent works are proposed to avoid checking all possible permutations of piece placements~\cite{gallagher2012jigsaw,son2014solving,chen2018new,mondal2013robust}. However, they mainly concentrate on the boundary information (four boundaries of each piece), whereas ignoring the semantic information that is useful for image understanding. Recently, some methods based on convolutional neural networks (CNN) have been introduced for predicting the permutations by neural optimization~\cite{sholomon2016dnn,dery2017neural} or the relative position of a fragment with respect to another~\cite{paumard2018image,paumard2020deepzzle,bridger2020solving}. Similar to previous methods, these algorithms do not consider the semantic content information and only rely on learning the position of each piece.

\begin{figure}[t]
    \centering
    \subfigcapskip=-5pt 
    \subfigure[Inputs]{
        \includegraphics[width=0.1166\textwidth]{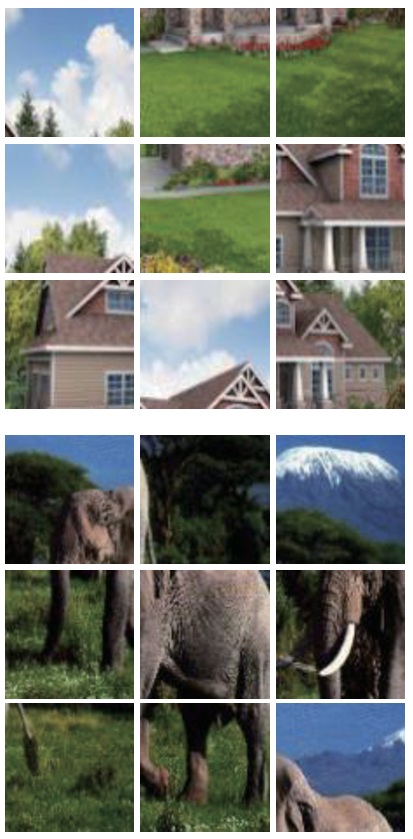}
    }
    \hspace{-0.45cm}
    \subfigure[Results of~\cite{paumard2020deepzzle}\hspace{0.2cm}]{
        \includegraphics[width=0.1194\textwidth]{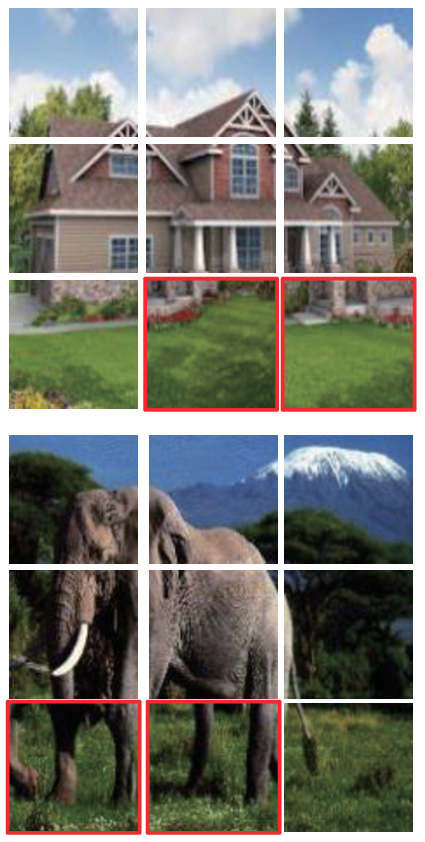}
    }
    \hspace{-0.447cm}
    \subfigure[Results of~\cite{huroyan2020solving}]{
        \includegraphics[width=0.1194\textwidth]{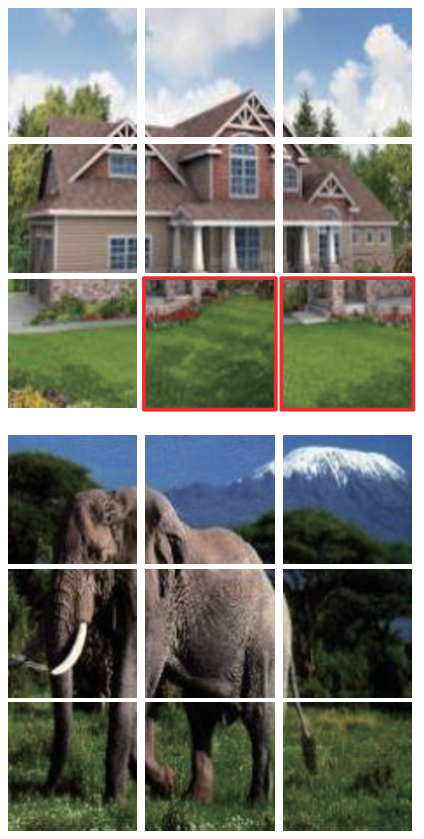}
    }
    \hspace{-0.435cm}
    \subfigure[Ours]{
        \includegraphics[width=0.1163\textwidth]{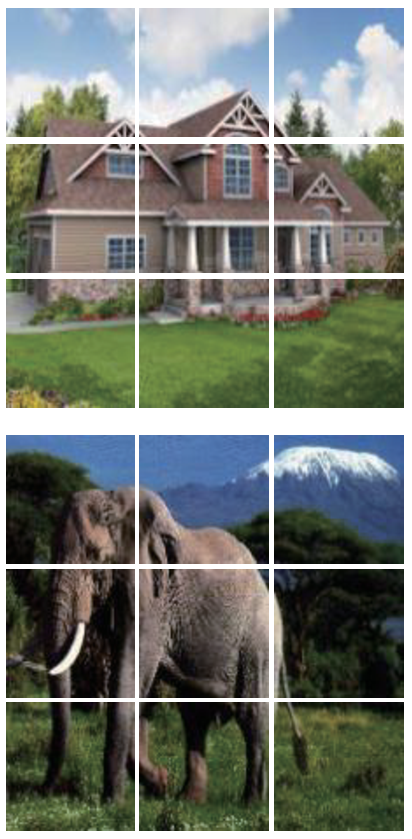}
    }    
    \caption{Comparisons with two recent jigsaw puzzle solvers. (a) Input images. (b) Results of Paumard~\emph{et al.}'s method~\cite{paumard2020deepzzle}. (c) Results of Huroyan~\emph{et al.}'s method~\cite{huroyan2020solving}. (d) Our results. Paumard~\emph{et al.}'s method~\cite{paumard2020deepzzle} cannot handle these two cases and Huroyan~\emph{et al.}'s method~\cite{huroyan2020solving} fails to solve the house example. The proposed JigsawGAN works well on these cases due to the concentration of boundary and semantic information simultaneously.}
    \label{fig:introduction}
\end{figure}

In this paper, we propose JigsawGAN, a GAN-based auxiliary learning pipeline that combines the boundary information of pieces and the semantic information of the global images to solve jigsaw puzzles, which can better simulate the procedure that how humans solve jigsaw puzzles. Auxiliary learning approaches aim to maximize the prediction of a primary task by supervising the model to additionally learn a secondary task~\cite{valada2018deep,lyu2020auxiliary}. In this work, we address the problem of permutation classification by simultaneously optimizing the GAN network as an auxiliary task. The proposed pipeline consists of two branches, a classification branch (Fig.~\ref{fig:pipeline} (b)) and a GAN branch (Fig.~\ref{fig:pipeline} (d), (e) and a discriminator). The classification branch classifies the jigsaw permutations, and the GAN branch recovers features to images with correct sequences. The two branches are connected by the encoder (Fig.~\ref{fig:pipeline} (a)) and a flow-based warp module (Fig.~\ref{fig:pipeline} (c)). 

The flow-based warp module is applied to warp encoder features to the predicted positions according to the classification results. Moreover, it contributes to the gradient back propagation from the GAN branch to the classification branch. The GAN branch consists of the decoder and the discriminator. The former one recovers the features to the images and the latter one aims to classify that an image belongs to the generated dataset or the real dataset. The discriminator can be regarded as the process that humans judge whether the reorganization of pieces is correct. The judgment process is essential to equip the pipeline with the capability of high-level image understanding, together with low-level boundary clues. The boundary loss and the GAN loss are applied for constraining the boundaries of pieces and distinguishing the semantic information, respectively. The JigsawGAN is mainly designed for solving $3 \times 3$ puzzles 
and existing algorithms for $3 \times 3$ puzzles can benefit from our method.
We introduce reference labels to guide the convergence of the classification network. The GAN branch can push the encoder to generate more informative features, thereby obtaining higher classification accuracy. 
Our method is able to improve the reorganization accuracy of existing methods if we select their results as our reference labels. Detailed descriptions and experiments 
are presented in Section~\ref{sec:ablation_improvement}.

Fig.~\ref{fig:introduction} shows the comparisons with two recent methods~\cite{paumard2020deepzzle,huroyan2020solving}. Paumard~\emph{et al.} proposed a CNN-based method to detect the neighbor pieces and applied the shortest path optimization to recover the images~\cite{paumard2020deepzzle}. Huroyan~\emph{et al.} applied the graph connection Laplacian algorithm to determine the boundary relationships~\cite{huroyan2020solving}.
Fig.~\ref{fig:introduction} exhibits the reassembled pieces according to the predicted labels. It includes two hard examples, in which three grassland pieces in the house example and two leg pieces in the elephant example can be easily confused.
Paumard~\emph{et al.}'s method~\cite{paumard2020deepzzle} cannot handle the two cases and Huroyan~\emph{et al.}'s method~\cite{huroyan2020solving} fails to solve the house example. In contrast, our method works well on these cases due to the concentration on the boundary information and semantic clues.

Overall, the main contributions are summarized as follows:
\begin{itemize}
\item We propose JigsawGAN, a GAN-based architecture to solve jigsaw puzzles, in which both the boundary information and semantic clues are fused for the inference. 
\item We introduce a flow-based warp module 
to reorganize feature pieces, which also ensures the gradient from the GAN branch can be back-propagated to the classification branch during the training process. 
\item We provide quantitative and qualitative comparisons with several typical jigsaw puzzle solvers to demonstrate the superiority of the proposed method.
\end{itemize}

\section{Related works}\label{sec:related_work}

Introduced by Freeman and Gardner~\cite{freeman1964apictorial}, the puzzle solving problems have been studied by many researchers, even though Demaine~\emph{et al.} discovered that puzzle assembling is NP-hard if the dissimilarity is unreliable~\cite{demaine2007jigsaw}. 

\subsection{Solving Square-piece Puzzles}
Most jigsaw solvers assume that the input includes equal-sized squared pieces. The first work was proposed by~\cite{cho2010probabilistic}, where a greedy algorithm and a benchmark were proposed. They presented a probabilistic solver to achieve approximated puzzle reconstruction. Pomeranz~\emph{et al.} evaluated some compatibility metrics and proposed a new compatibility metric to predict the probability that two given parts are neighbors~\cite{pomeranz2011fully}. Gallagher~\emph{et al.} divided the squared jigsaw puzzle problems into three types~\cite{gallagher2012jigsaw}. `Type 1' puzzle means the orientation of each jigsaw piece is known, and only the location of each piece is unknown. `Type 2' puzzle is defined as a non-overlapping square-piece jigsaw puzzle with unknown dimension, unknown piece rotation, and unknown piece position. Meanwhile, the global geometry and position of every jigsaw piece in `Type 3' puzzle are known, and only the orientation of each piece is unknown. Gallagher~\emph{et al.} solved `Type 1' and `Type 2' puzzle problems through the minimum spanning tree (MST) algorithm constrained by geometric consistency between pieces. A Markov Random Field (MRF)-based algorithm was also proposed to solve the `Type 3' puzzle. 

\begin{figure*}[t]
   \centering
   \includegraphics[width=0.98\textwidth]{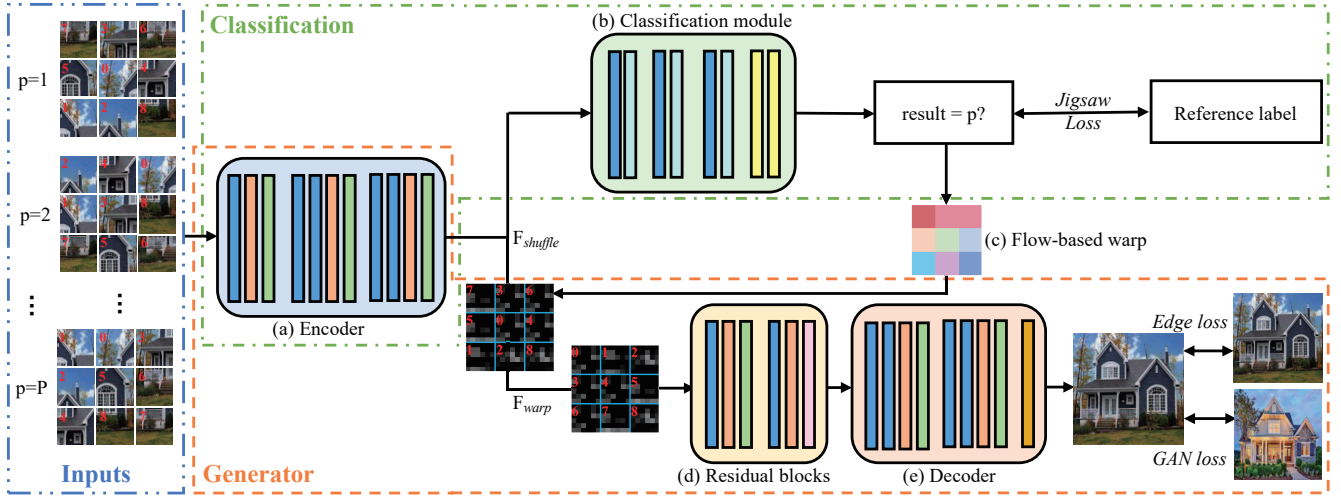}\\
   \caption{The pipeline exhibits the architecture of the proposed JigsawGAN, which consists of two streams that operate different functions. \textbf{The top stream} classifies the jigsaw permutations. In turn,
   \textbf{the bottom stream} recovers features to images. The two streams are connected by the encoder and a flow-based warp module. 
   The encoder extracts features of input shuffled pieces. The shuffled features are then combined, called $F_{shuffle}$, to be fed to the classification module to predict the jigsaw permutation. The classification results provides information to generate flow fields that can reassemble the combined feature $F_{shuffle}$ to a new combined feature $F_{warp}$. The warped feature $F_{warp}$ is then sent to the residual blocks and the decoder to recover natural image. 
   }
   \label{fig:pipeline}
\end{figure*}

Some subsequent works were proposed to solve the `Type 1' puzzle problem.
Yu~\emph{et al.} applied linear programming to exploit all pairwise matches, and computed the position of each component~\cite{yu2015solving}.
There are also some methods aimed to solve the complicated `Type 2' puzzle problem. Son~\emph{et al.} introduced loop constraints for assembling non-overlapping jigsaw puzzles where the rotation and the position of each piece are unknown~\cite{son2014solving}. Huroyan~\emph{et al.}  utilized graph connection Laplacian to recover the rotations of the pieces when both shuffles and rotations are unknown~\cite{huroyan2020solving}. 
Some variants were investigated subsequently, including handling puzzles with missing pieces~\cite{paikin2015solving} and eroded boundaries~\cite{paumard2020deepzzle}. 
Paikin~\emph{et al.} calculated dissimilarity between pieces and then proposed a greedy algorithm for handling puzzles of unknown size and with missing entries~\cite{paikin2015solving}. Bridger~\emph{et al.} proposed a GAN-based architecture to recover the eroded regions and reassembled the images using greedy algorithms~\cite{bridger2020solving}.  
Our method concentrates on solving the `Type 1' non-overlapping square-piece jigsaw puzzles.

Current DL-based methods are mainly designed for solving 9 pieces with strong robustness compared to traditional algorithms. Dery~\emph{et al.}~\cite{dery2017neural} applied a pre-trained VGG model as feature extractor and corrected the order using a pointer network~\cite{vinyals2015pointer}. Paumard~\emph{et al.} utilized a CNN-based method to detect the neighbor pieces and used the shortest path optimization to recover the eroded images~\cite{paumard2020deepzzle}. 
These methods obey the conventional rule to solve jigsaw puzzles, which consists of two steps. The first step is applying the neural network to determine the relationships of pieces, and the second step is using optimization techniques to reorganize the pieces. Their performances are limited by the first step due to the aperture problem based only on the local pieces. We propose a GAN-based approach to include global semantic information apart from local pieces for a better accuracy. 

\begin{table*}[t]
\centering
\caption{Layer configurations of the architecture. $B$ represents the batch size, $H_p$ and $W_p$ are the height and width of input pieces. The encoder extracts features of input shuffled pieces. The shuffled features are then combined ($F_{shuffle}$) with size $H_f$ and $W_f$. The warped feature $F_{warp}$ is sent to the residual blocks and the decoder to recover natural image. 
}
\begin{minipage}[c]{0.45\textwidth}
\subtable[Encoder]{
\centering
\begin{tabular}{ccccc}
\hline
\multicolumn{5}{c}{Input: {[}$B \times n \times n$, 3, $H_p$, $W_p${]}}                                                            \\ \hline
\multicolumn{3}{c|}{Conv}                      & \multicolumn{1}{c|}{BN}      & \multirow{2}{*}{Activation} \\ \cline{1-4}
Kernel & Stride & \multicolumn{1}{c|}{Channel} & \multicolumn{1}{c|}{Channel} &                             \\ \hline
7      & 1      & \multicolumn{1}{c|}{64}      & \multicolumn{1}{c|}{64}      & ReLU                        \\
3      & 2      & \multicolumn{1}{c|}{128}     & \multicolumn{1}{c|}{-}       & -                           \\
3      & 1      & \multicolumn{1}{c|}{128}     & \multicolumn{1}{c|}{128}     & ReLU                        \\
3      & 2      & \multicolumn{1}{c|}{256}     & \multicolumn{1}{c|}{-}       & -                           \\
3      & 1      & \multicolumn{1}{c|}{256}     & \multicolumn{1}{c|}{256}     & ReLU                        \\ \hline
\multicolumn{5}{c}{$F_{shuffle}$: {[}$B$, 256, $H_f$, $W_f${]}}                                                             \\ \hline
\end{tabular}
}
\end{minipage}
\begin{minipage}[c]{0.5\textwidth}
\centering
\subtable[Classification module]{
\centering
\begin{tabular}{cccccc}
\hline
\multicolumn{6}{c}{$F_{shuffle}$: {[}$B$, 256, $H_f$, $W_f${]}}                                                                                                                      \\ \hline
\multicolumn{3}{c|}{Conv}                      & \multicolumn{1}{l|}{\multirow{2}{*}{Activation}} & \multicolumn{1}{c|}{Max pooling} & FC      \\ \cline{1-3}
Kernel & Stride & \multicolumn{1}{c|}{Channel} & \multicolumn{1}{l|}{}                            & \multicolumn{1}{c|}{Stride}      & Channel \\ \hline
3      & 2      & \multicolumn{1}{c|}{256}     & \multicolumn{1}{c|}{ReLU}                        & \multicolumn{1}{c|}{2}           & -       \\
3      & 2      & \multicolumn{1}{c|}{384}     & \multicolumn{1}{c|}{ReLU}                        & \multicolumn{1}{c|}{2}           & -       \\
3      & 2      & \multicolumn{1}{c|}{384}     & \multicolumn{1}{c|}{ReLU}                        & \multicolumn{1}{c|}{2}           & -       \\
-      & -      & \multicolumn{1}{c|}{-}       & \multicolumn{1}{c|}{-}                           & \multicolumn{1}{c|}{-}           & 4096    \\
-      & -      & \multicolumn{1}{c|}{-}       & \multicolumn{1}{c|}{-}                           & \multicolumn{1}{c|}{-}           & $P$     \\ \hline
\multicolumn{6}{c}{Classification output: $B \times P$}                                                                                                \\ \hline
\end{tabular}
}
\end{minipage}
\begin{minipage}[c]{0.45\textwidth}
\subtable[Residual blocks and decoder]{
\centering
\begin{tabular}{ccccc}
\hline
\multicolumn{5}{c}{$F_{warp}$: {[}B, 256, $H_f$, $W_f${]}}                                                                                   \\ \hline
\multicolumn{3}{c|}{Conv}                      & \multicolumn{1}{c|}{BN}      & \multirow{2}{*}{Activation} \\ \cline{1-4}
Kernel & Stride & \multicolumn{1}{c|}{Channel} & \multicolumn{1}{c|}{Channel} &                             \\ \hline
3      & 2      & \multicolumn{1}{c|}{256}     & \multicolumn{1}{c|}{256}     & ReLU                        \\
3      & 1      & \multicolumn{1}{c|}{256}     & \multicolumn{1}{c|}{256}     & ES                          \\ \hline
3      & 1/2    & \multicolumn{1}{c|}{128}     & \multicolumn{1}{c|}{-}       & -                           \\
3      & 1      & \multicolumn{1}{c|}{128}     & \multicolumn{1}{c|}{128}     & ReLU                        \\
3      & 1/2    & \multicolumn{1}{c|}{64}      & \multicolumn{1}{c|}{-}       & -                           \\
3      & 1      & \multicolumn{1}{c|}{64}      & \multicolumn{1}{c|}{64}      & ReLU                        \\
7      & 1      & \multicolumn{1}{c|}{3}       & \multicolumn{1}{c|}{-}       & -                           \\ \hline
\multicolumn{5}{c}{Generator output: {[}$B$, 3, $H$, $W${]}}                                                     \\ \hline
\end{tabular}
}
\end{minipage}
\begin{minipage}[c]{0.5\textwidth}
\centering
\subtable[Discriminator]{
\centering
\begin{tabular}{ccccc}
\hline
\multicolumn{5}{c}{Input: {[}$B$, 3, $H$, $W${]}}                                                                 \\ \hline
\multicolumn{3}{c|}{Conv}                      & \multicolumn{1}{c|}{BN}      & \multirow{2}{*}{Activation} \\ \cline{1-4}
Kernel & Stride & \multicolumn{1}{c|}{Channel} & \multicolumn{1}{c|}{Channel} &                             \\ \hline
3      & 1      & \multicolumn{1}{c|}{32}      & \multicolumn{1}{c|}{-}       & LReLU                       \\
3      & 2      & \multicolumn{1}{c|}{64}      & \multicolumn{1}{c|}{-}       & LReLU                       \\
3      & 1      & \multicolumn{1}{c|}{64}      & \multicolumn{1}{c|}{64}      & LReLU                       \\
3      & 2      & \multicolumn{1}{c|}{128}     & \multicolumn{1}{c|}{-}       & LReLU                       \\
3      & 1      & \multicolumn{1}{c|}{128}     & \multicolumn{1}{c|}{128}     & LReLU                       \\
3      & 1      & \multicolumn{1}{c|}{256}     & \multicolumn{1}{c|}{256}     & LReLU                       \\
3      & 1      & \multicolumn{1}{c|}{1}       & \multicolumn{1}{c|}{-}       & -                           \\ \hline
\multicolumn{5}{c}{Output: {[}$B$, 1, $H/4$, $W/4${]}}                                                            \\ \hline
\end{tabular}
}
\end{minipage}
\label{tab:layer_config}
\end{table*}

\subsection{Pre-training Jigsaw Puzzle Solvers}
Numerous self-supervised methods consider solving jigsaw puzzles as pre-text tasks. This learning strategy is a recent variation of the unsupervised learning theme that transfers the pre-trained network parameters on jigsaw puzzle tasks to other visual recognition tasks~\cite{doersch2015unsupervised}.
These methods assume that a rich universal representation has been captured in the pre-trained model, which is useful to be fined-tuned with the task-specific data using various strategies. Noroozi~\emph{et al.} introduced a context-free network (CFN) to separate the pieces in the convolutional process. Its main architecture focused on a subset of possible permutations including all the image tiles and solved a classification problem ~\cite{noroozi2016unsupervised}. Santa~\emph{et al.} proposed to handle the whole set by approximating the permutation matrix and solving a bi-level optimization problem to recover the right ordering~\cite{santa2017deeppermnet}. The above methods tackled the problem by dealing with the separate pieces and then finding a way to recombine them. Carlucci~\emph{et al.} proposed JiGen to train a jigsaw classifier and a object classifier simultaneously. They focused on domain generalization tasks by considering that the jigsaw puzzle solver can improve semantic understanding~\cite{carlucci2019domain}. Du~\emph{et al.} combined the jigsaw puzzle and the progressive training to optimize the fine-grained classification by learning which granularities are the most discriminative and how to fuse information cross multi-granularity~\cite{du2020fine}. However, solving jigsaw puzzle tasks in these methods are supervised, which depend heavily on the training data. 
In this paper, we propose a GAN-based auxiliary learning method to solve jigsaw puzzles, where the ground truth of jigsaw puzzle task is unavailable. Auxiliary learning in our method means that we formulate the permutation classification as the primary task with the secondary goal is to optimize the GAN network.

\section{Method}

We propose a GAN-based architecture for solving the jigsaw puzzles with unpaired datasets. The architecture includes two streams operating different functions. The first stream defines the jigsaw puzzle solving as a classification task to judge which permutation the shuffled input belongs to. The second stream is composed of the generator $G$ and the discriminator $D$. The generator $G$ learns the mapping function between different domains, while the discriminator $D$ aims to optimize the generator by distinguishing target domain images from the generated ones.
Specifically, we generate a wide diversity of shuffled images ${\{ {x_i}\} _{i = 1,...,N}} \in X$ as the source domain data, and a collection of natural images ${\{ {y_j}\} _{j = 1,...,M}} \in Y$, as the target domain data. The data distributions of the two domains are denoted as $x\sim{p_{{\rm{data}}}}(x)$ and $y\sim{p_{{\rm{data}}}}(y)$, respectively.

\subsection{Network Architecture}\label{sec:architecture}
We present the classification network $C$, generator $G$ in Fig.~\ref{fig:pipeline}. where $C$, $G$ and the discriminator $D$ are simultaneously optimized during the training process. 
Their detailed layer configurations are shown in Table~\ref{tab:layer_config}. The inputs are obtained by decomposing the source images into $n \times n$ pieces, which are then shuffled and reassigned to one of the ${n^2}$ grid positions to generate ${n^2}!$ combinations. ${n^2}!$ is a very large number and scarcely possible to be considered as classification tasks if $n>2$, for example, $(3^2)! = 362, 880$. Generally, $P$ elements are selected from $n^2!$ combinations according to the maximal Hamming distance~\cite{noroozi2016unsupervised}, and we assign an index to each entry. In our implementation, solving the jigsaw puzzle is considered as a classification task and each permutation corresponds to a classification label. The number of permutations indicates the classification categories. The ultimate target is to predict correct permutation of the shuffled input image.

The shuffled input image is first divided into $n \times n$ pieces, and the pieces are sent to the network in parallel. We choose the discrete pieces as inputs in order to prevent the cross-influence between the boundaries of pieces. The encoder (Fig.~\ref{fig:pipeline} (a)) extracts features of these shuffled pieces. The shuffled features are then combined, called $F_{shuffle}$, to be sent to the classification module (Fig.~\ref{fig:pipeline} (b)) to predict the jigsaw permutation. The classification module is constrained by the pseudo-labels, called reference labels. During the training, we find that an unsupervised classification network converges only with difficulty. To solve the problem, we generate the reference labels to guide $C$. The classification results provide information to generate flow fields (Fig.~\ref{fig:pipeline} (c)) that can reassemble the combined feature $F_{shuffle}$ to a new combined feature $F_{warp}$. The warped feature $F_{warp}$ is then sent to the residual blocks (Fig.~\ref{fig:pipeline} (d)) and the decoder (Fig.~\ref{fig:pipeline} (e)) to recover natural images. The decoder can recover a perfect reorganized image if $F_{warp}$ is reassembled with the correct order according to the classification label.

When the real permutation labels and the correct natural images are unavailable, we adopt the GAN architecture to achieve the unsupervised optimization. The vanilla GAN consists of two components, a generator $G$ and a discriminator $D$, where $G$ is responsible for capturing the data distribution while $D$ tries to distinguish whether a sample comes from the real data or the generator. This framework corresponds to a min-max two-player game, and introduces a powerful way to estimate the target distribution. The GAN branch provides global constraints to enable the network to concentrate on semantic clues. The decoder and the discriminator as well as extra losses enable the encoder to generate more informative features and further improve the classification network.

\subsubsection*{Classification Network} 
The classification network consists of two parts: the convolutional blocks (the encoder) and the classification module. The encoder extracts useful signals for downstream transformation. The classification module aims to distinguish different permutations, which includes convolutional layers, max pooling layers and fully connected layers.

\subsubsection*{Generator} 
The generator network is composed of three parts: the encoder, eight residual blocks and the decoder. The encoder of the classification network also extracts features for the generator. Afterward, eight residual blocks with identical layouts are adopted to construct the content and the manifold features. The decoder consists of two identical transposed convolutional blocks and a final convolutional layer.

\subsubsection*{Discriminator} 
The discriminator network complements the generator and aims to classify each image as real or fake. The discriminator network includes several convolutional blocks. Such a simple discriminator uses fewer parameters and can work on images of arbitrary sizes.

\begin{figure}[t]
   \centering
   \includegraphics[width=0.945\linewidth]{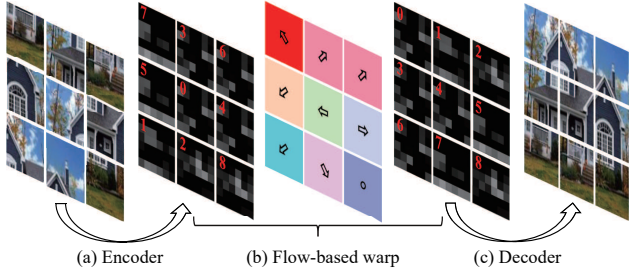}\\
   \caption{The flow-based warp module can warp the shuffled features to ordered features according to the classification results.}
   \label{fig:flow_warp}
\end{figure}

\subsection{Flow-based Warp Module}\label{sec:flow_warp}

The GAN branch cannot recover the extracted features directly because the features are shuffled. We introduce a flow-based warp module to reassemble the shuffled encoder features and ensure the gradient can be back-propagated correctly during the feature recombination. In detail, we define the flow fields to represent dense pixel correspondences between the shuffled encoder features ($F_{shuffle}$) and the warpped features ($F_{warp}$). The optical flows are constructed according to the predicted labels, which are used to warp $F_{shuffle}$ into $F_{warp}$. The size of the flow-based warp module is defined as $H_f \times W_f \times 2$, where the first $H_f \times W_f$ channel controls the horizontal shift of the features and the last $H_f \times W_f$ channel calculates the vertical shift of the features. The discrete encoder features are combined before reassembling by the flow-based warp module. $H_f$ and $W_f$ are the height and width of combined features. The flow-based warp module not only rearranges the shuffled features but also guarantees the gradient back propagation. Note that, some shift operations can also recombine the shuffled features, but block the gradient back propagation. As a result, the semantic information from the GAN branch cannot be delivered to the classification branch, thereby resulting in lower classification accuracy.

The decoder can recover perfect images if the classification labels are accurate, as shown in Fig.~\ref{fig:flow_warp}. Some wrong classification can be tolerated when the network is initialized because the GAN optimization can correct the mistakes gradually. GAN network can detect the reorganization error and rectify it according to the semantic information acquired from the real dataset. The correct information will be transferred to the encoder and the classification module through the gradient back propagation, and can guide the classification network to classify the input correctly on following iterations. 

\subsection{Loss Function}\label{sec:loss}
We design our objective function to include the following three losses: (1) the jigsaw loss ${L_{jigsaw}}(C)$, which optimizes the classification network to recognize the correct permutations; (2) the adversarial loss ${L_{{\rm{GAN}}}}(G,D)$, which drives the generator network to achieve the desired transformation; (3) the boundary loss ${L_{boundary}}(G,D)$, which pushes the decoder of $G$ to recover clear images, and further constrains the encoder to generate useful features that are helpful for the classification network. The full objective function is:
\begin{equation}\label{eq:loss}
\small
L(G,D,C) = {L_{jigsaw}}(C) + {L_{{\rm{GAN}}}}(G,D) + {L_{boundary}}(G,D) 
\end{equation}

\subsubsection{Jigsaw Loss}\label{sec:jig_loss}

We consider the jigsaw puzzle solving as a classification task and $P$ is the classification category. Kullback-Leibler (KL) divergence can measure the similarity between the predicted distribution and the target distribution. The KL divergence becomes smaller when two permutations tend to be similar. We aim to minimize the KL divergence between the reassembled result ($p_{predict}$) and the ground truth ($p_{real}$), which can be described as follows: 
\begin{equation}\label{eq:minmax}
\arg \mathop {\min } KL(p_{predict}, p_{real})
\end{equation}

A simple classification network is first proposed to recognize the correct permutation. 
We minimize the following jigsaw loss to optimize the classification network:
\begin{equation}\label{eq:jigsaw_loss1}
{L_{jigsaw}}(C) = {\mathbb{E}_{x\sim{p_{{\rm{data}}}}(x)}}[CE(C(x),p)],
\end{equation}
where $p$ is the probability distribution of the real data, $C(x)$ is the probability distribution of the predicted data which indicates the probability that the result belongs to each category. $p$ and $C(x)$ are defined as matrix with size $B \times P$, where $B$ is the batch size. CE is cross-entropy loss which is defined as:
\begin{equation}\label{eq:CE_loss1}
CE(C(x),p)=-\sum p \cdot log(C(x))
\end{equation}

\begin{figure}[t]
   \centering
   \includegraphics[width=0.96\linewidth]{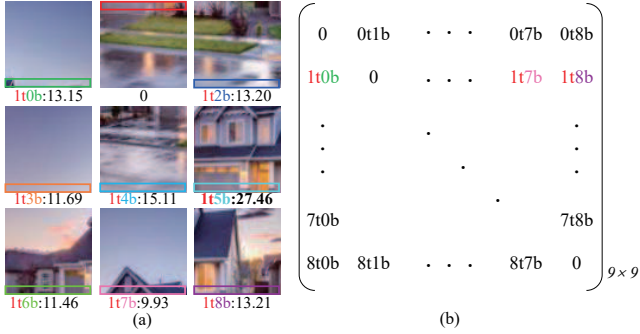}\\
   \caption{In this case, we calculate the PSNR values between the top boundary of the top-middle piece and the bottom boundaries of other pieces. Corresponding PSNR values are displayed in (a). These PSNR values compose the second row of the 9 $\times$ 9 top-bottom relationship metric, as shown in (b).
   }
   \label{fig:edge_psnr}
\end{figure}

However, the real jigsaw labels are unavailable for unsupervised tasks. Directly training $C$ without jigsaw labels is impossible. The classification network tends to assign labels randomly, which achieves 30-40\% classification accuracy for three-classification tasks and 20-30\% for four-classification tasks according to our experiments. Note that, three-classification tasks mean the dataset contains three categories and the model will predict the most likely category that one input belongs to. Pseudo-labels are introduced to achieve better classification performance. The pseudo-labels, called reference labels, are generated according to the shuffled input images. The reference labels can constrain $C$ when permutation indexes are unavailable.
We apply the following five steps to obtain the reference labels:

\begin{enumerate}
    \item Cutting four boundaries from each piece, and the width of the boundary is determined by a hyper-parameter $pix$. 
    \item Calculating the PSNR values between the top boundary of a specific piece and bottom boundaries of other pieces to obtain a vector with size $1 \times {n^2}$. Then, a ${n^2} \times {n^2}$ matrix can be obtained when computes all top-bottom relationships, as shown in Fig.~\ref{fig:edge_psnr}.
    \item Applying the same way to get another ${n^2} \times {n^2}$ matrix to indicate the left-right relationships.
    \item There are $n \times (n-1)$ correct top-bottom boundary pairs and $n \times (n-1)$ correct left-right boundary pairs, so we adopt a greedy algorithm~\cite{paumard2018jigsaw} to select $n \times (n-1)$ maximum values from two matrices, respectively.
    \item A minimum spanning tree (MST) algorithm~\cite{gallagher2012jigsaw} is then applied to assemble the pieces, and the reference permutation ${p_{ref}}$ is returned.
\end{enumerate}

\begin{table}[]
\centering
\caption{Experiments to show the influence of the hyper-parameter $pix$.}
\label{tab:ablation_edge_pixel}
\begin{tabular}{cccc}
\toprule
         & $pix$ = 3 & $pix$ = 2 & $pix$ = 1 \\
\midrule
PSNR & 20.1 & 21.5 & 24.7  \\
Accuracy & 56.5\% & 60.3\% & 66.4\%  \\
\bottomrule
\end{tabular}
\vspace{-0.1cm}
\end{table}

Empirically, $pix$ is set to 1, which means we cut a row of pixels of each boundary. Table~\ref{tab:ablation_edge_pixel} shows the corresponding PSNR values and reorganization accuracy when we select different $pix$. A larger $pix$ may cut a wider boundary strip. However, the redundant information could decrease the PSNR values and the reorganization accuracy. We select $pix$ = 1 when we obtain the boundary strips, which are more sensitive to the boundary discontinuity.

After the reference labels are obtained, focal (FL) loss~\cite{lin2017focal} is applied as the jigsaw loss. FL loss ensures the minimization of the KL divergence whilst increasing the entropy of the predicted distribution, which can prevent the model from becoming overconfident. FL loss can down-weight easy examples and focus training on hard negatives. We replace the CE loss conventionally used with the FL loss to improve the network calibration. FL loss is defined as:
\begin{equation}\label{eq:FL_loss1}
FL(C(x),p)=-\sum (1-C(x)\cdot p)^\gamma \cdot log(C(x))
\end{equation}
where $(1-C(x)\cdot p)^\gamma$ is a modulating factor to optimize the imbalance of the dataset and $\gamma$ is a tunable focusing parameter that smoothly adjusts the rate at which easy examples are down-weighted. $\gamma$ is set to 2 in our implementations. Finally, the jigsaw loss can be described as follows:
\begin{equation}\label{eq:jigsaw_loss2}
{L_{jigsaw}}(C) = {\mathbb{E}_{x\sim{p_{{\rm{data}}}}(x)}}[FL(C(x), p_{ref})]
\end{equation}

\subsubsection{Adversarial Loss}

The performance of the classification module $C$ is limited by the reference labels. After the encoder module generating distinctive features, we add a decoder module to recover the features to natural images. We do not train an autoencoder network that directly recovers the shuffled inputs because it is useless for the encoder to generate more informative features. The proposed flow-based warp module is able to warp the shuffled features ($F_{shuffle}$) to warpped features ($F_{warp}$) according to the classification results. The reorganized features $F_{warp}$ are then sent to the GAN branch to recover the images. The GAN branch is auxiliary to the classification branch and trained to help the encoder module generate useful features that are aware of the semantic information, and further improve the classification performance.

As in classic GAN networks, the adversarial loss is used to constrain the results of $G$ to look like target domain images. In our task, adversarial loss pushes $G$ to generate natural images in the absence of corresponding ground truth. Meanwhile, $D$ aims to distinguish whether a given image belongs to the synthesized or the real target set, which tries to classify an image into two categories: the generated image $G(x)$ and the real image $y$, as formulated in Eq.~\ref{eq:gan_loss}.
\begin{equation}\label{eq:gan_loss}
\begin{aligned}
{L_{{\rm{GAN}}}} & = {\mathbb{E}_{y\sim{p_{{\rm{data}}}}(y)}}\big[\log D(y)\big] \\
& + {\mathbb{E}_{x\sim{p_{{\rm{data}}}}(x)}}\big[\log (1 - D(G(x))\big]
\end{aligned}
\end{equation}

\begin{figure}[t]
   \centering
   \includegraphics[width=0.96\linewidth]{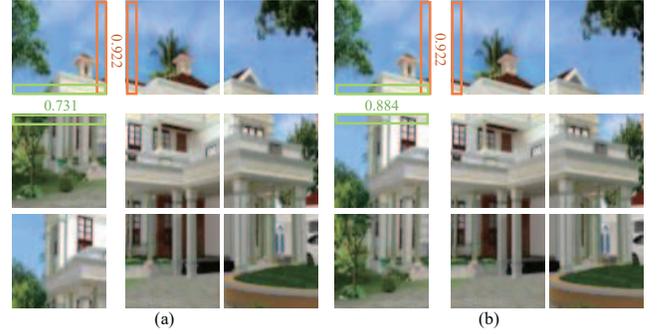}\\
   \caption{(a) An incorrect reorganized result that may confuse the discriminator. (b) The correct reassemble result. The numbers indicate the SSIM values between two adjacent boundaries.}
   \label{fig:gan_error}
\end{figure}

\begin{figure*}[t]
   \centering
   \includegraphics[width=0.965\textwidth]{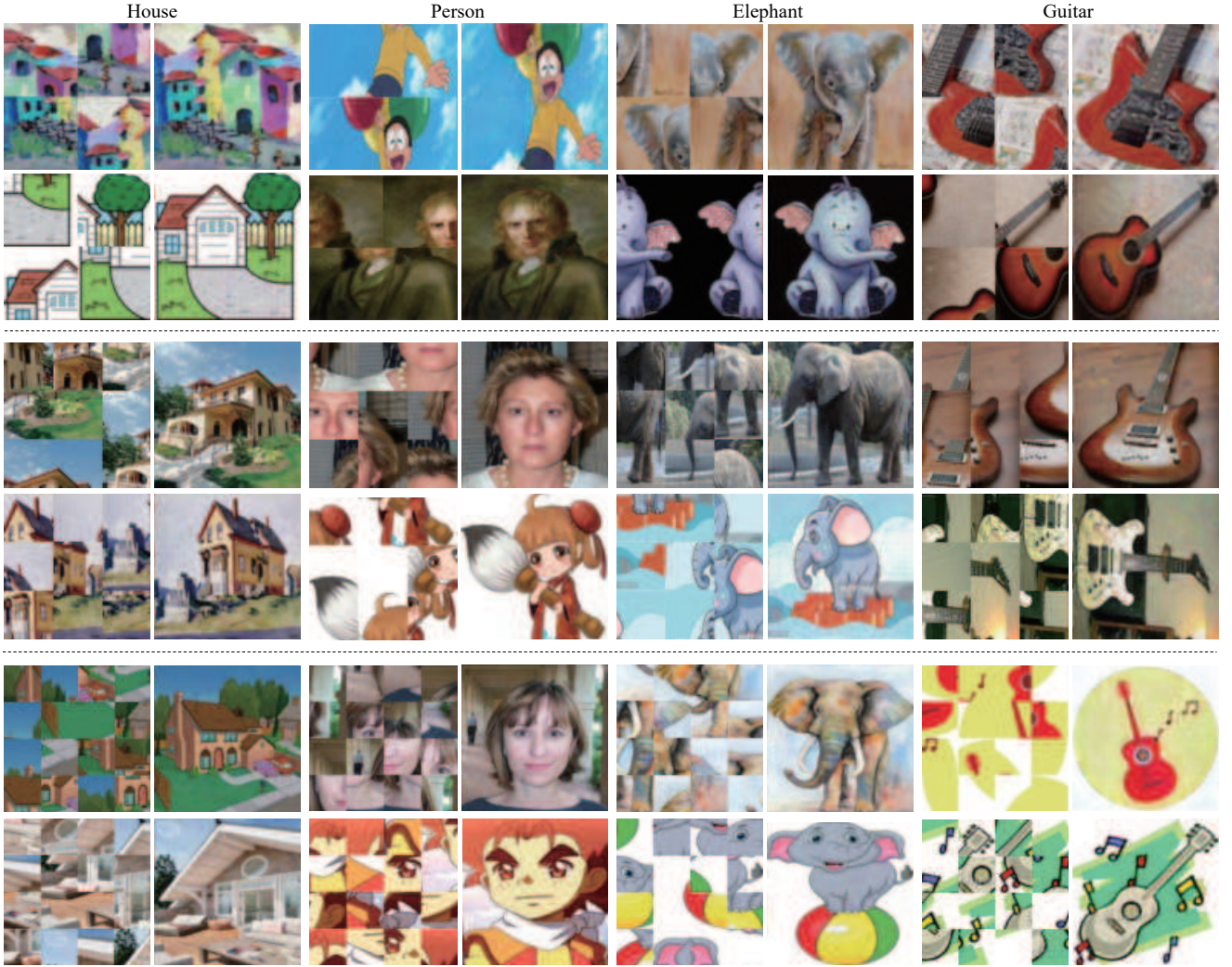}\\
   \caption{Examples generated by the proposed JigsawGAN. The top two rows show the results when the hyper-parameter $n = 2$, the middle two rows present the results when the hyper-parameter $n = 3$, and the bottom two rows are results when $n = 4$. Different styles can be efficiently reassembled by JigsawGAN.}
   \label{fig:reorganization1}
\vspace{-0.2cm}
\end{figure*}

\subsubsection{Boundary Loss}

The adversarial loss ensures that the generated image looks similar to the target domain. It is inadequate for a smooth transaction, especially at the boundaries of pieces. For example, in Fig.~\ref{fig:gan_error} (a), sorting adjacent top-bottom pieces in house images mistakenly may confuse the discriminator, such that the discriminator considers the imperfect image as correct image and cannot work as we desired. Therefore, it is essential to enforce a more strict constraint to guarantee the semantic consistency of the generated images. To achieve this, we add penalties at the boundaries of pieces to punish the incorrect combination. We first cut the decoder result into $n \times n$ pieces and calculate the SSIM of adjacent boundaries. For example, as for $3 \times 3$ puzzle, the top-left piece needs to calculate the SSIM of its right boundary with the left boundary of the top-middle piece (orange boxes in Fig.~\ref{fig:gan_error}). Similarly, the SSIM value between its bottom boundary and the top boundary of the middle-left piece is computed (green boxes in Fig.~\ref{fig:gan_error}). We obtain $n \times (n-1)$ values to represent the top-bottom relationships and $n \times (n-1)$ values to indicate the left-right relationships. Two average scores of these two relationships 
are calculated to design the boundary loss, noted as $SSIM_{tb}$ and $SSIM_{lr}$, respectively. As such, the boundary loss is defined as:
\begin{equation}\label{eq:edge_loss}
\begin{aligned}
{L_{boundary}}(G, D) &= {\mathbb{E}_{x\sim{p_{{\rm{data}}}}(x)}}[(1 - SSIM_{tb} (G(x)))] \\
& +  {\mathbb{E}_{x\sim{p_{{\rm{data}}}}(x)}}[(1 - SSIM_{lr} (G(x)))]
\end{aligned}
\end{equation}

Matching boundaries generate large SSIM values. The boundary loss provides more penalties to wrong placements.

\section{Experiments}

We first illustrate the datasets and implementation details in Section~\ref{sec:implementation} and then conduct comprehensive experiments to verify the effectiveness of the proposed method, including quantitative comparisons (Section~\ref{sec:quantitative}), qualitative comparisons (Section~\ref{sec:qualitative}), computational times (Section~\ref{sec:complexity}) and ablation studies (Section~\ref{sec:ablation}). Specifically, we first compare our method with several representative jigsaw puzzle works, including a classic jigsaw puzzle solver~\cite{gallagher2012jigsaw}, a linear programming-based method~\cite{yu2015solving}, a loop constraints-based method~\cite{son2014solving}, a graph connection Laplacian-based method~\cite{huroyan2020solving}, and then compare with a recent DL-based method which relies on the shortest path optimization~\cite{paumard2020deepzzle}. Some comparison methods are designed for handling the `Type 1' and `Type 2' puzzles simultaneously, which can be easily applied for our `Type 1' task for solving $3 \times 3$ pieces. Next, ablation studies are performed to illustrate the importance of different components.

\begin{table*}[t]
\centering
\caption{Quantitative comparisons with several representative jigsaw puzzle solvers in terms of reorganization accuracy.}
\label{tab:quantitative}
\begin{tabular}{ccccccc}
\toprule
& Gallagher2012~\cite{gallagher2012jigsaw} & Son2014~\cite{son2014solving} & Yu2015~\cite{yu2015solving} & Huroyan2020~\cite{huroyan2020solving} & Paumard2020~\cite{paumard2020deepzzle} & Ours  \\
\midrule
House & 64.5\%  & 70.3\%   &  73.8\%    & 76.4\%  & 61.2\% & \textbf{80.2\%} \\
Person & 63.8\%  & 68.2\%   &  70.6\%    & 73.7\%  & 63.6\% & \textbf{79.9\%} \\
Guitar & 59.4\%  & 64.5\%   &  68.3\%    & 71.1\%  & 60.4\% & \textbf{77.4\%} \\
Elephant & 60.5\%  & 66\%   &  71.4\%    & 73.1\%  & 58.2\% & \textbf{78.5\%} \\
\midrule
Mean & 62.1\%  & 67.3\%   &  71.0\%    & 74.6\%  & 60.9\% & \textbf{79.0\%} \\
\bottomrule
\end{tabular}
\vspace{-0.2cm}
\end{table*}

\subsection{Datasets and Implementation Details}\label{sec:implementation}

\subsubsection{Data Collection}
The dataset includes 7,639 images, among which 5,156 images originate from the PACS dataset~\cite{li2017deeper} and 2,483 images are from our own collection. The PACS dataset includes many pictures with similar contents, so we gathered images from movies or Internet that are more distinguishable to increase the diversity. Our dataset covers 4 object categories (House, Person, Elephant and Guitar), and each of them can be divided into 4 domains (Photo, Art paintings, Cartoon and Sketches). Each category is divided into three subsets: a jigsaw set (40\%) to generate the inputs, a real dataset (40\%) to help the discriminator to distinguish the generated images and real images, and a test set (20\%) to evaluate the performance of different methods. They are randomly selected from the overall dataset. The shuffled fragments are then prepared according to~\cite{carlucci2019domain}: the input image is divided into $n \times n$ pieces and the size of each piece is set to 24 $\times$ 24. The permutation $P$ is set to 1,000 in our implementation. The hyper-parameter $n$ and the permutation $P$ are important factors for the network performance. We perform ablation studies on these two factors in Section~\ref{sec:ablation_n_and_p}.

\subsubsection{Training Details}
We implement JigsawGAN in PyTorch, and all the experiments are performed on an NVIDIA RTX 2080Ti GPU with 100 epochs. For each iteration, every input image is divided into $n \times n$ pieces, which are sent to the network in parallel. We choose the discrete pieces as inputs in order to prevent the cross-influence between the boundaries of pieces. Current architecture can handle a maximum of 16 pieces. If we want to deal with images with more pieces,  the classification network should be deepened with more convolutional blocks to get a larger respective field.
Adam optimizer is applied with the learning rate of $2.0{\rm{ \times 1}}{{\rm{0}}^{{\rm{ - 4}}}}$ for the generator, the discriminator and the classification network. 
The entire training process costs 4 hours on average.

\subsection{Quantitative Comparisons}\label{sec:quantitative}
Table~\ref{tab:quantitative} lists the reorganization accuracy of the aforementioned methods and the JigsawGAN when $n=3$. The reorganization accuracy means the proportion of correct reassembled images in all test images, and the scores reported in Table~\ref{tab:quantitative} are the average results over the test sets. Some comparison methods~\cite{gallagher2012jigsaw,son2014solving,yu2015solving,huroyan2020solving} obey the 
basic rules to solve jigsaw puzzles: (1) detecting boundaries to determine the relationship between the pieces, and (2) applying different optimization methods to reassemble the images. Gallagher~\emph{et al.} applied the Mahalanobis Gradient Compatibility (MGC) measurement to determine the pieces' relationship and a greedy algorithm to reorganize the pieces~\cite{gallagher2012jigsaw}. A considerable improvement for~\cite{gallagher2012jigsaw} was proposed by~\cite{son2014solving}, by adding loop constraints to the piece reorganization process. Yu~\emph{et al.} combined the advantages of greedy methods and loop propagation algorithms to introduce a linear programming-based solver~\cite{yu2015solving}. Huroyan~\emph{et al.} applied the graph connection Laplacian to better understand the reconstruction mechanism~\cite{huroyan2020solving}. However, these methods are limited by the boundary detection step, which will lead to the randomness of their results. Paumard~\emph{et al.}~\cite{paumard2020deepzzle} applied the shortest path optimization algorithm to reorganize the pieces, whose boundary relationship is predicted by a neural network. Their pruning strategy for the shortest path algorithm may affect the performance. Compared to these methods, our JigsawGAN achieves the best scores, which improves the performance against~\cite{gallagher2012jigsaw,son2014solving,paumard2020deepzzle} with a relatively large margin and has more than 4 point improvements compared with~\cite{yu2015solving,huroyan2020solving}.  

\begin{figure*}[t]
   \centering
   \includegraphics[width=0.988\textwidth]{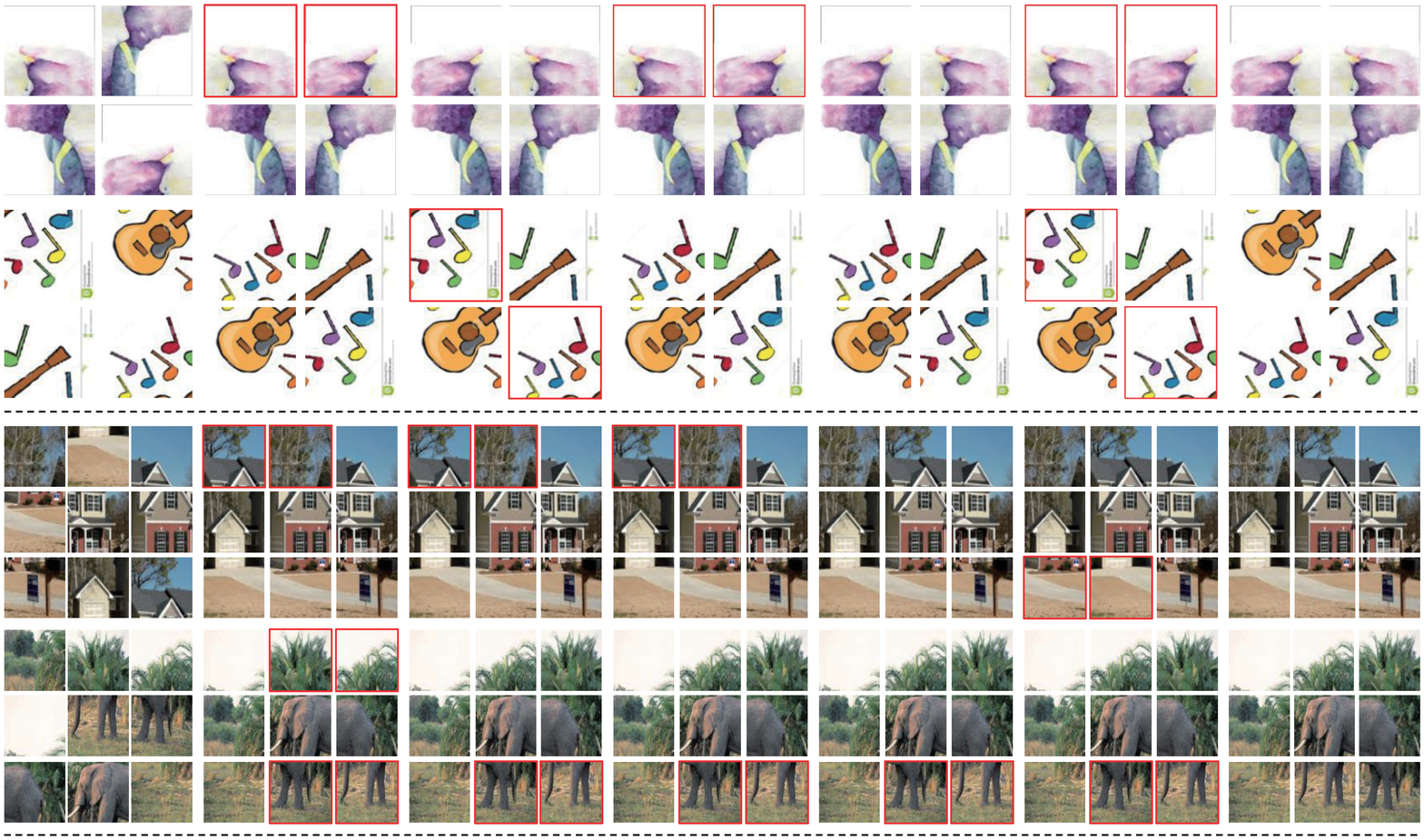}\\
   \label{fig:fig:reorganization1}
   \vspace{-0.475cm}
\end{figure*}
\begin{figure*}[t]
    \centering
    \subfigcapskip=-4pt
    \subfigure[Inputs]{
        \includegraphics[width=0.1313\textwidth]{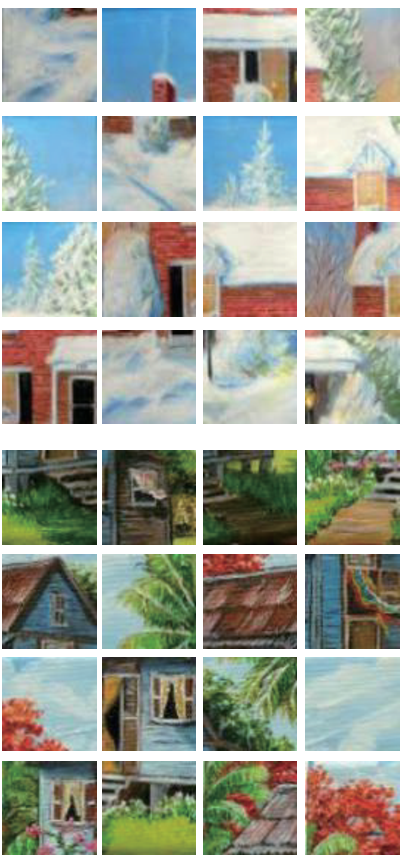}
    }
    \hspace{-0.38cm}
    \subfigure[Results of~\cite{gallagher2012jigsaw}]{
        \includegraphics[width=0.1352\textwidth]{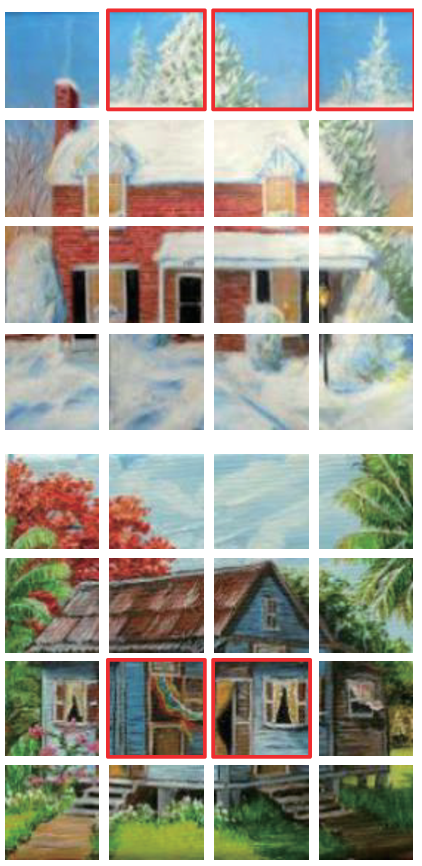}
    }
    \hspace{-0.34cm}
    \subfigure[Results of~\cite{son2014solving}]{
        \includegraphics[width=0.137\textwidth]{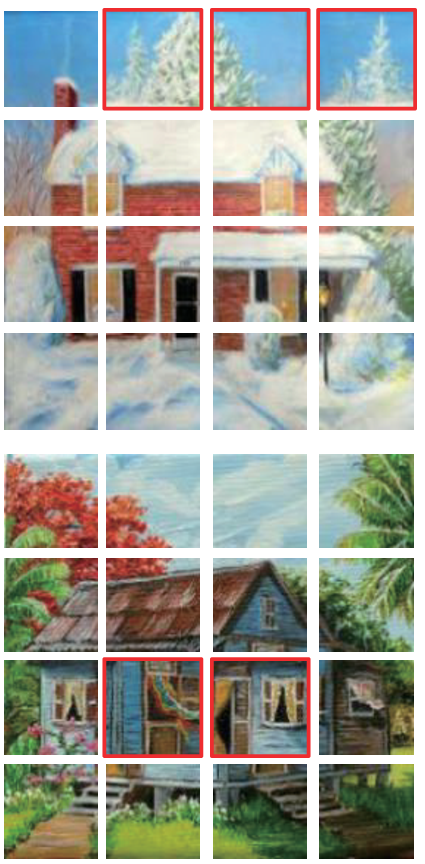}
    }
    \hspace{-0.37cm}
    \subfigure[Results of~\cite{yu2015solving}]{
        \includegraphics[width=0.1378\textwidth]{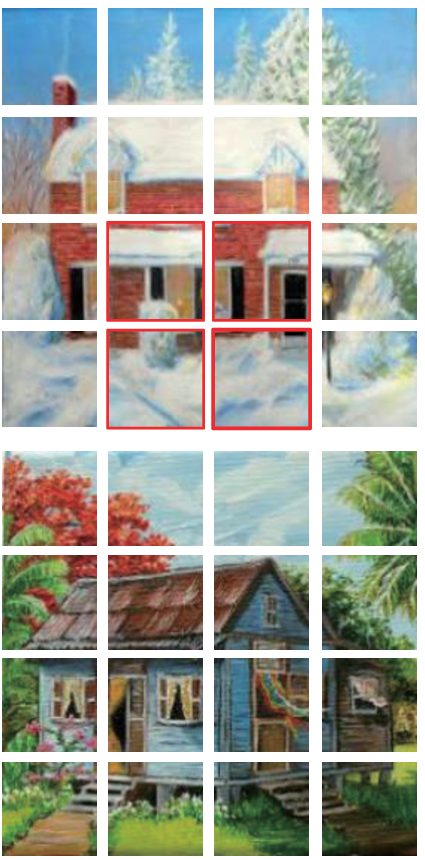}
    }
    \hspace{-0.37cm}
    \subfigure[Results of~\cite{huroyan2020solving}]{
        \includegraphics[width=0.1385\textwidth]{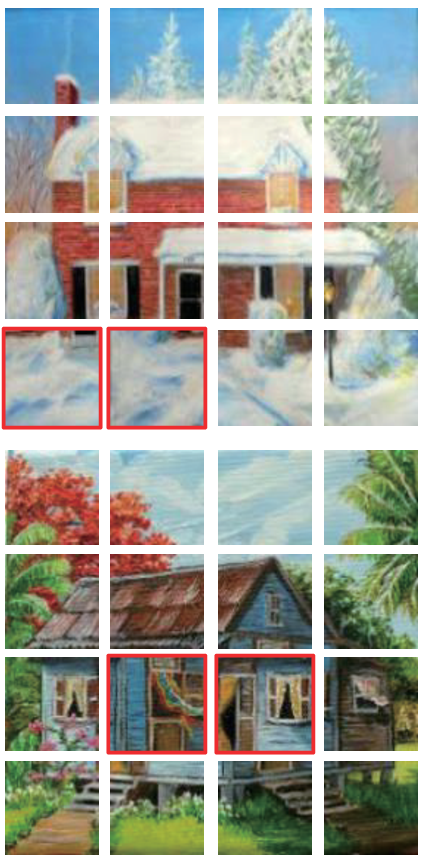}
    }
    \hspace{-0.40cm}
    \subfigure[Results of~\cite{paumard2020deepzzle}]{
        \includegraphics[width=0.1393\textwidth]{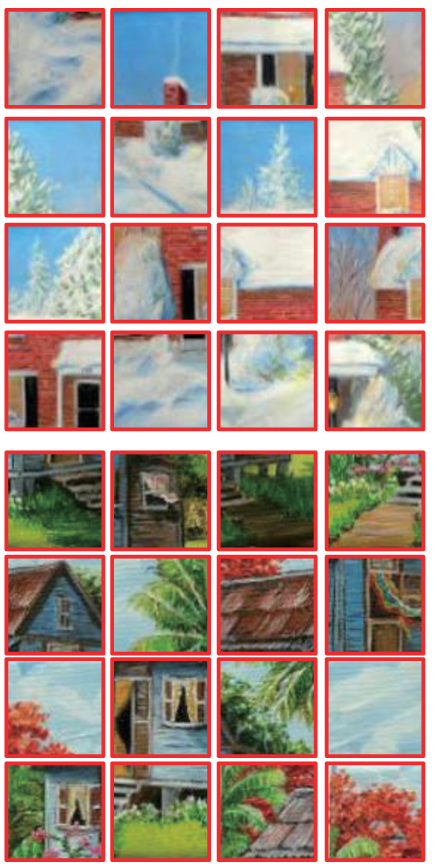}
    }
    \hspace{-0.38cm}
    \subfigure[Ours]{
        \includegraphics[width=0.134\textwidth]{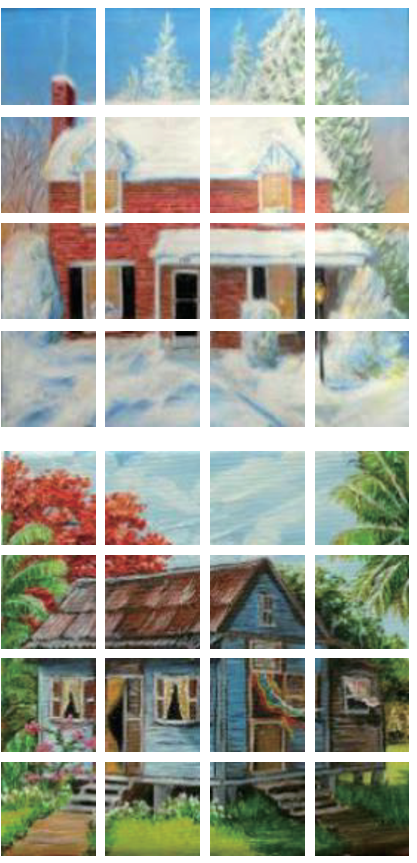}
    }
    \caption{Qualitative comparisons with Gallagher~\emph{et al.}'s method~\cite{gallagher2012jigsaw}, Son~\emph{et al.}'s method~\cite{son2014solving}, Yu~\emph{et al.}'s method~\cite{yu2015solving}, Huroyan~\emph{et al.}'s method~\cite{huroyan2020solving} and Paumard~\emph{et al.}'s method~\cite{paumard2020deepzzle}. The top two rows show the results when the hyper-parameter $n = 2$, the middle two rows present the results when $n = 3$, and the bottom two rows are results when $n = 4$. The proposed method assembles both the boundary and the semantic information in order to generate more accurate results. The results of~\cite{paumard2020deepzzle} are the same as the inputs when $n = 4$ because their method can only handle puzzles when $n \le 3$.}
    \label{fig:reorganization2}
\vspace{-0.2cm}
\end{figure*}

\begin{table*}[!ht]
\centering
\caption{Computational times of JigsawGAN and the comparison methods when solving $3\times3$ puzzles.}
\label{tab:complexity}
\begin{tabular}{ccccccc}
\toprule
Methods & Gallagher2012~\cite{gallagher2012jigsaw} & Son2014~\cite{son2014solving} & Yu2015~\cite{yu2015solving} & Huroyan2020~\cite{huroyan2020solving} & Paumard2020~\cite{paumard2020deepzzle} & Ours \\
\midrule
Time ($s$) & 0.274     & 1.121 & 0.525 & 0.496   & 3.054   & 0.293 \\
\bottomrule
\end{tabular}
\end{table*}

\subsection{Qualitative Comparisons}\label{sec:qualitative}
In this section, we first show some reorganization results generated by JigsawGAN in Fig.~\ref{fig:reorganization1}. Our method can reassemble the inputs and generate high-quality results. Then, qualitative comparisons with aforementioned methods are presented in Fig.~\ref{fig:reorganization2}, which shows the reassembled pieces according to corresponding predicted labels. The top two rows show the results when the hyper-parameter $n = 2$, the middle two rows present the results when $n = 3$, and the bottom two rows are results when $n = 4$. In the first elephant example, the top two pieces can be easily confused if the methods do not consider the semantic content information. Results of~\cite{gallagher2012jigsaw,yu2015solving} and~\cite{paumard2020deepzzle} fail to solve the case. The guitar example has weak boundary constraints between the pieces, which leads to the failure of~\cite{son2014solving} and~\cite{paumard2020deepzzle}. The strategy of Son~\emph{et al.} aims to recover the complete shape from pieces based on a dissimilarity metric~\cite{son2014solving}. The two musical note pieces have a small contribution to construct the guitar structure and confuse Son~\emph{et al.}'s method. Paumard~\emph{et al.} reassemble pieces with boundary erosion, which directly determine the relationships of pieces through a neural network and ignore the useful boundary information~\cite{paumard2020deepzzle}. In the first house example, results of~\cite{gallagher2012jigsaw,son2014solving} and~\cite{yu2015solving} fail to solve the tree and roof pieces, while the result of~\cite{paumard2020deepzzle} cannot distinguish two road pieces. In the second elephant example, they all fail to discriminate the leg pieces. 
We noted that puzzles with low percentages of recovery by these algorithms have large portions of pieces with the same uniform texture and color. The global semantic information and the strict boundary detection are indispensable to obtain better reassemble results. The failure of the last two house examples of other methods when $n = 4$ is caused by the similar texture and the unobvious boundary. Note that, the results of~\cite{paumard2020deepzzle} are the same as the inputs when $n = 4$ because their method can only handle puzzles when $n \le 3$.
In comparison, by considering the global information (GAN loss) and the boundary information (boundary loss) simultaneously, our method recovers these cases well.

\subsection{Computational Times}\label{sec:complexity}

Computing efficiency is also important for evaluating the reassemble performance. We conduct comparisons of computational times with the aforementioned methods. The results for solving $3\times3$ puzzles on the test set are reported in Table~\ref{tab:complexity}. Gallagher~\emph{et al.}'s method~\cite{gallagher2012jigsaw} is the fastest algorithm and our method is a bit slower than~\cite{gallagher2012jigsaw}. However, introducing a useful network with a little more running time is advisable in exchange for better results. Methods~\cite{yu2015solving} and~\cite{huroyan2020solving} are slower than JigsawGAN due to the linear programming algorithm and the optimized iteration, respectively. As for~\cite{son2014solving}, most of the time is spent in the pairwise matching, the unoptimized merging, the trimming and the filling steps. Paumard~\emph{et al.}'s method~\cite{paumard2020deepzzle} is the slowest due to the complicated network architecture and the reassemble graph. Although our method takes approximately 4 hours on the training process, the PSNR-based reference label and the GAN network are not used in the test process, which take less running time on the test set. The simple yet effective classification network can provide satisfactory reassemble results. 

\subsection{Ablation Studies}\label{sec:ablation}

\subsubsection{Ablation Study of Loss Terms}
We perform the ablation study on the variants of the loss function to understand how these main modules contribute to the final results. Table~\ref{tab:ablation_loss} displays the ablation results, which demonstrates that each component contributes to the objective function.
We have illustrated the importance of the jigsaw loss ${L_{jigsaw}}$ in Section~\ref{sec:jig_loss}. Directly training the classification network without the jigsaw loss is impossible. The network tends to assign results randomly, achieving 30-40\% classification accuracy for three-class tasks and 20-30\% for four-class tasks. Table~\ref{tab:ablation_loss} further shows the importance of the adversarial loss ${L_{{\rm{GAN}}}}$ and the boundary loss ${L_{boundary}}$. As shown in Table~\ref{tab:ablation_loss}, removing ${L_{{\rm{GAN}}}}$ degrades the results substantially, so as removing the ${L_{boundary}}$. The third column shows the reorganization accuracy without ${L_{{\rm{GAN}}}}$,  which indicates the discriminator and corresponding loss are removed. The accuracy of without ${L_{{\rm{GAN}}}}$ is apparently inferior to the final result. The fourth column displays the result without ${L_{boundary}}$, which means the decoder is only constrained by the adversarial loss. The result is also not as good as our final result. We also conduct the experiment that removes both ${L_{{\rm{GAN}}}}$ and ${L_{boundary}}$, which represents the GAN branch is not involved, where the final classification accuracy only depends on the reference labels ${p_{ref}}$. The result in the second column is worse than the final result. We conclude that all three loss terms are critical.

\begin{table}[t]
\centering
\caption{Ablation experiments of different loss terms.}
\label{tab:ablation_loss}
\begin{tabular}{ccccc}
\toprule
         & ${p_{ref}}$  & w/o. ${L_{{\rm{GAN}}}}$ & w/o. ${L_{boundary}}$ & Ours   \\
\midrule
Accuracy & 66.4\% & 68.1\% & 75.8\% & 79.0\% \\
\bottomrule
\end{tabular}
\end{table}

\begin{table}[t]
\centering
\caption{Ablation experiments of different datasets. P, H, G and E means the Person, House, Guitar and Elephant datasets, respectively. `H.G.E and P' represents that we train our network on House, Guitar and Elephant datasets, and then test on the Person dataset.}
\label{tab:ablation_dataset}
\begin{tabular}{cccc}
\toprule
         & P and P & H.P.G.E and P & H.G.E and P\\
Accuracy & 79.9\% & 72.4\% & 66.9\%\\
\midrule
        & H and H & H.P.G.E and H & P.G.E and H\\
Accuracy & 80.2\% & 74.6\% & 69.7\%\\
\midrule
         & G and G & H.P.G.E and G & H.P.E and G\\
Accuracy & 77.4\% & 69.8\% & 63.1\%\\
\midrule
         & E and E & H.P.G.E and E & H.P.G and E \\
Accuracy & 78.5\% & 72.2\% & 65.5\%\\
\bottomrule
\end{tabular}
\end{table}

\subsubsection{Ablation Study of Different Training Sets}
This ablation study aims to demonstrate the effectiveness of the semantic information provided by the GAN branch. In Table~\ref{tab:ablation_dataset}, `P' is the `Person' dataset, `H' means the `House' dataset, `G' represents the `Guitar' dataset and `E' indicates the `Elephant' dataset. For each item in Table~\ref{tab:ablation_dataset}, the capital letter before `and' means we train our network on corresponding datasets, and the capital letter after `and' means we test the model on corresponding datasets. For example, `H.G.E and P' represents that we train the network on House, Guitar and Elephant datasets, and then test on the Person dataset.  Table~\ref{tab:ablation_dataset} exhibits four groups of experiments according to different datasets. 

\begin{table}[t]
\centering
\caption{The reorganization accuracy when select different $n$ and $P$.}
\label{tab:ablation_nandp}
\begin{tabular}{cccc}
\toprule
  & $P = 10$     & $P = 100$    & $P = 1000$   \\
\midrule
$n = 2$ & 91.9\% & 87.0\% & 83.2\% \\
$n = 3$ & 85.6\% & 83.3\% & 79.0\% \\
\bottomrule
\end{tabular}
\end{table}

\begin{table*}[t]
\centering
\caption{The improvement when select different methods as our reference labels.}
\label{tab:ablation_improvement}
\begin{tabular}{cccccc}
\toprule
            & Gallagher2012~\cite{gallagher2012jigsaw} & Son2014~\cite{son2014solving} & Yu2015~\cite{yu2015solving} & Huroyan2020~\cite{huroyan2020solving} & Paumard2020~\cite{paumard2020deepzzle}\\
\midrule
Original accuracy      & 62.1\% & 67.3\% & 71.0\% & 74.6\% & 60.9\% \\
Ours        & 75.6\% & 78.4\% & 79.2\% & 81.0\% & 69.4\% \\
Improvements & +13.5\% & +11.1\% & +7.8\% & +6.4\% & +8.5\%  \\
\bottomrule
\end{tabular}
\end{table*}

We first select `P' as the test dataset and set `P', `H.P.G.E' and `H.G.E' as the training dataset, respectively. When the training and the test dataset belong to the same category, the network can learn the semantic information and obtain the best performance (79.9\%). Then, if we set all datasets as the training set, the performance is inferior to the result when the network is trained on a single dataset because the mixed semantic information will influence the judgment of the discriminator (72.4\%). To explore the effectiveness of the semantic information deeply, we further conduct the experiment when selecting `H.G.E' as the training dataset while the test set is `P'. The result shows that the network can hardly obtain the correct semantic information if the training set and the test set were less correlated. Therefore, the result (66.9\%) is similar to the result when we removing the GAN branch (66.4\%). For example, as for the `Person' category, if the generated image exchanges the arm and leg pieces, the GAN network can judge that it is irrational according to the semantic information acquired from the real `Person' dataset, and give more penalties to the case. Otherwise, if the real dataset contains images coming from other datasets, the composition of `Person' cannot be detected accurately. The trend of other three datasets is the same as the Person dataset (row 3 to row 8), which demonstrate the semantic information provided by the GAN branch is important.

\subsubsection{Ablation Study of Grids and Permutations}\label{sec:ablation_n_and_p}
The selection of different $n$ and $P$ have significant influences on the reorganization accuracy. As shown in Table~\ref{tab:ablation_nandp}, increasing the permutation $P$ obviously degrades the accuracy, so as increasing the hyper-parameter $n$. Increasing $P$ makes the classification more complicated, while increasing $n$ improves the difficulty of detecting the relationships of pieces. With the increase of $P$, the permutations become close to each other and their features tend to be similar. It is challenging for the classification network to recognize them correctly, and therefore affecting the classification accuracy.

\subsubsection{Ablation Study of Reference Labels}\label{sec:ablation_improvement}
The proposed method can be considered as an improvement technique of existing methods, which can improve the reorganization accuracy of them if we select their results as our reference labels.
As described in Section~\ref{sec:related_work}, the jigsaw puzzle tasks can be specifically divided into three categories: (1) conventional methods that solve hundreds of pieces; (2) DL-based methods which generally handle $3 \times 3$ pieces; (3) self-supervised learning methods that consider solving jigsaw puzzles as pre-text tasks.  The first category methods concentrate on obtaining better performance in detecting the neighbor pieces, whereas ignoring the accuracy of reassembling the images correctly. If we consider assigning each piece to the correct place as a perfect reconstruction, the proportion of correct reconstruction images in all test images of their methods only occupies 50\%-60\% approximately. The proposed method belongs to the second category and considers solving jigsaw puzzles as a classification task. We define \emph{homogeneous algorithms} to indicate the methods that can be applied to solve $3 \times 3$ puzzles, including the second category method~\cite{paumard2020deepzzle} and some first category methods ~\cite{gallagher2012jigsaw,son2014solving,yu2015solving,huroyan2020solving}. The proposed JigsawGAN is compatible with homogeneous methods to create considerable improvement, which is benefited from the GAN branch and corresponding losses.

Table~\ref{tab:ablation_improvement} shows the improvements of aforementioned methods if we choose their results as our reference labels, respectively. As for our results, higher reference label accuracy leads to higher reorganization accuracy because the reference labels can better guide the network to learn useful information, such as achieving 81.0\% accuracy for method~\cite{huroyan2020solving}. The improvements of~\cite{gallagher2012jigsaw} and~\cite{son2014solving} are significant, while the improvements of~\cite{yu2015solving} and~\cite{huroyan2020solving} are relatively smaller because there is an upper bound of the network. As for reference labels with lower accuracy, the GAN branch can promote them significantly. Moreover, it is reasonable that the improvement for~\cite{paumard2020deepzzle} is small because~\cite{paumard2020deepzzle} is mainly designed for their proposed dataset with eroded boundaries. Overall, no matter how the reference labels are obtained, it can be promoted by utilizing our proposed architecture.

\subsection{Discussion}
Solving the jigsaw puzzle with deep learning methods is a developing research area, especially when the inputs include many pieces. Conventional DL-based methods concentrate on solving puzzles with $3 \times 3$ pieces. The upper limit of the proposed architecture is determined by the respective field. Current architecture can handle a maximum of 16 pieces. If we want to deal with images with more pieces, the classification network should be deepened with more convolutional blocks to get a larger respective field. Moreover, solving non-square puzzles is still a challenging research problem. The proposed method in the current version is constrained by the reference label and the boundary loss, which are designed to optimize straight boundaries. Applying a metric designed for matching non-square pieces to compute the boundary loss and update the reference label may be useful for solving arbitrary-size puzzles. We consider them as future works.

\section{Conclusion}
We have proposed JigsawGAN, a GAN-based auxiliary learning method for solving jigsaw puzzles when the prior knowledge of the initial images is unavailable. 
The proposed method can apply the boundary information of pieces and the semantic information of generated images to solve jigsaw puzzles more accurately. The architecture contains the classification branch and the GAN branch, which are connected by an encoder and a flow-based warp module. The GAN branch drives the encoder to generate more information features and further improves the classification branch. GAN loss and a novel boundary loss are introduced to constrain the network to focus on the semantic information and the boundary information, respectively. We have conducted comprehensive experiments to demonstrate the effectiveness of JigsawGAN. 


\ifCLASSOPTIONcaptionsoff
  \newpage
\fi


\end{document}